%% file: ms.tex
\documentclass[10pt,twocolumn,letterpaper]{article}

\usepackage{cvpr}              % To produce the CAMERA-READY version
\usepackage{graphicx}
\usepackage{amsmath}
\usepackage{amssymb}
\usepackage{booktabs}

\usepackage[svgnames,table]{xcolor}
\usepackage{enumitem}
\usepackage{adjustbox}

\usepackage{array}
\usepackage{subcaption}
\usepackage{multirow}
\usepackage{rotating}

\newcolumntype{x}[1]{>{\centering\arraybackslash}p{#1}}

\usepackage[pagebackref,breaklinks,colorlinks]{hyperref}

\usepackage[capitalize]{cleveref}
\crefname{section}{Sec.}{Secs.}
\Crefname{section}{Section}{Sections}
\Crefname{table}{Table}{Tables}
\crefname{table}{Tab.}{Tabs.}

\def\cvprPaperID{5871} % *** Enter the CVPR Paper ID here
\def\confName{CVPR}
\def\confYear{2023}

\begin{document}

\title{Tunable Convolutions with Parametric Multi-Loss Optimization}

\author{
    Matteo Maggioni, Thomas Tanay, Francesca Babiloni, Steven McDonagh, Ale\v{s} Leonardis \\
    {Huawei Noah's Ark Lab}\\
    {\tt\footnotesize \{matteo.maggioni, thomas.tanay, francesca.babiloni, steven.mcdonagh, ales.leonardis\}@huawei.com}
    % {\tt\footnotesize firstname.lastname@huawei.com}
}

\maketitle

%%%%%%%%% ABSTRACT
\begin{abstract}
Behavior of neural networks is irremediably determined by the specific loss and data used during training. However it is often desirable to tune the model at inference time based on external factors such as preferences of the user or dynamic characteristics of the data. This is especially important to balance the perception-distortion trade-off of ill-posed image-to-image translation tasks. In this work, we propose to optimize a parametric tunable convolutional layer, which includes a number of different kernels, using a parametric multi-loss, which includes an equal number of objectives. Our key insight is to use a shared set of parameters to dynamically interpolate both the objectives and the kernels. During training, these parameters are sampled at random to explicitly optimize all possible combinations of objectives and consequently disentangle their effect into the corresponding kernels. During inference, these parameters become interactive inputs of the model hence enabling reliable and consistent control over the model behavior. Extensive experimental results demonstrate that our tunable convolutions effectively work as a drop-in replacement for traditional convolutions in existing neural networks at virtually no extra computational cost, outperforming state-of-the-art control strategies in a wide range of applications; including image denoising, deblurring, super-resolution, and style transfer.
\end{abstract}

%%%%%%%%% BODY TEXT
\section{Introduction}
\label{sec:intro}

Neural networks are commonly trained by optimizing a set of learnable weights against a pre-defined loss function, often composed of multiple competing objectives which are delicately balanced together to capture complex behaviors from the data. 
Specifically, in vision, and in image restoration in particular, many problems are ill-posed, \ie admit a potentially infinite number of valid solutions~\cite{hadamard1923lectures}. Thus, selecting an appropriate loss function is necessary to constrain neural networks to a specific inference behavior~\cite{zhao2017loss,mustafa2022task}. However, any individual and fixed loss defined empirically before training is inherently incapable of generating optimal results for any possible input~\cite{raychaudhuri2022multi}. A classic example is the difficulty in finding a good balance for the perception-distortion trade-off~\cite{zhao2017loss,blau2018perception}, as shown in the illustrative example of Fig.~\ref{fig:teaser}.
The solution to this problem is to design a mechanism to reliably control (\ie \emph{tune}) neural networks at inference time. This comes with several advantages, namely providing a flexible behavior without the need to retrain the model, correcting failure cases on the fly, and balancing competing objectives according to user preference.

\input{figures/1_teaser}

Existing approaches to control neural networks, commonly based on weights~\cite{wang2018esrgan,wang2019dni} or feature~\cite{wang2019cfsnet,zhang2020dynet} modulation, are fundamentally limited to consider only two objectives, and furthermore require the addition of a new set of layers or parameters for every additional loss considered. Different approaches, specific to image restoration tasks, first train a network conditioned on the true degradation parameter of the image, \eg noise standard deviation or blur size, and then, at inference time, propose to interact with these parameters to modulate the effects of the restoration~\cite{he2020cresmd,tseng2021differentiable,jiang2021jpeg}. However this leads the network to an undefined state when asked to operate in regimes corresponding to combinations of input and parameters unseen during training~\cite{kim2021cirn}.

In this work, we introduce a novel framework to reliably and consistently tune model behavior at inference time. We propose a parametric dynamic layer, called tunable convolution, consisting in $p$ individual kernels (and biases) which we optimize using a parametric dynamic multi-loss, consisting in $p$ individual objectives. Different parameters can be used to obtain different combinations of kernels and objectives by linear interpolation. The key insight of our work is to establish an explicit link between the $p$ kernels and objectives using a \emph{shared} set of $p$ parameters. Specifically, during training, these parameters are randomly sampled to explicitly optimize the complete loss landscape identified by all combinations of the $p$ objectives. As a result, during inference, each individual objective is disentangled into a different kernel, and thus its influence can be controlled by interacting with the corresponding parameter of the tunable convolution. In contrast to previous approaches, our strategy is capable of handling an arbitrary number of objectives, and by explicitly optimizing all their intermediate combinations, it allows to tune the overall network behavior in a predictable and intuitive fashion. Furthermore, our tunable layer can be used as a drop-in replacement for standard layers in existing neural networks with negligible difference in computational cost.

\noindent In summary the main contributions of our work are:
\setlist{nolistsep}
\begin{itemize}[noitemsep]
    \item A novel plug-and-play tunable convolution capable to reliably control neural networks through the use of interactive parameters;
    \item A unique parametric multi-loss optimization strategy dictating how tunable convolution should be optimized to disentangle the different objectives into the different tunable kernels;
    \item Extensive experimental validation across several image-to-image translation tasks demonstrating state-of-the-art performance for tunable inference.
\end{itemize}

\section{Related Works}
\label{sec:related}

In this section, we position our contribution and review relevant work. We put a particular emphasis on image-to-image translation and inverse imaging (\eg denoising and super-resolution) as these problems are clear use-cases for dynamic and controllable networks due to their ill-posedness.

\textbf{Dynamic Non-Interactive Networks.}
Contemporary examples of dynamic models~\cite{han2021dynamic} introduce dynamic traits by adjusting either model structure or parameters adaptively to the input at inference time. Common strategies to implement dynamic networks use attention or auxiliary learnable modules to modulate convolutional weights~\cite{dai2017deformable,su2019pixel,yang2019condconv,chen2020dynamic,zhang2020dynet}, recalibrate features~\cite{hu2018squeeze,park2019spade,li2019selective}, or even directly predicting weights specific to the given input~\cite{ha2016hypernetworks,jia2016dynamic,mildenhall2018kpn,xia2020basis}. These methods strengthen representation power of the network through the construction of dynamic and discriminative features, however do not offer interactive controls over their dynamicity.

\textbf{Parametric Non-Interactive Networks.}
The use of auxiliary input information is a strategy commonly used to condition the behavior of neural networks and to improve their performance~\cite{park2019spade}. In the context of image restoration, this information typically represents one or more degradation parameters in the input image, \eg noise standard deviation, blur size, or JPEG compression level, which is then provided to the network as additional input channels~\cite{zhang2018ffdnet,gharbi2016deep,mildenhall2018kpn}, or as parameters to modulate convolutional weights~\cite{luo2021functional}. This approach is proven to be an effective way to improve performance of a variety of image restoration tasks, including image denoising, super-resolution, joint denoising \& demosaicing, and JPEG deblocking~\cite{gharbi2016deep,luo2021functional}, however it is not explicitly designed to control the underlying network at inference time.

\textbf{Interactive Single-Objective Networks.}
More recently, a number of works have explored the use of external degradation parameters not only as conditional information to the network, but also as a way to enable interactive image restoration. 
A recent example is based on feature modulation inspired by squeeze-and-excitation~\cite{hu2018squeeze} where the excitation weights are generated by a fully-connected layer whose inputs are the external degradation parameters~\cite{he2020cresmd}. A similar modulation strategy is also employed in~\cite{jiang2021jpeg} for JPEG enhancement where a fully-connected layer controlling the quality factor is used to generate affine transformation weights to modulate features in the decoder stage of the network. In these works, the network is optimized using a fixed loss to maximize restoration performance conditioned on the \emph{true} parameters. Then, at inference time, the authors propose to use of these parameters to change the model behavior, \eg using a noise standard deviation lower than the real one to reduce the denoising strength. However, in these cases the network is requested to operate outside its training distribution, and --unsurprisingly-- is likely to generate suboptimal results which often contain significant artifacts, as also observed in~\cite{kim2021cirn}.
Finally, in a somewhat different spirit, alternative approaches use external parameters to modulate the noise standard deviation~\cite{kwon2021display}, or to build task-specific networks via channel pruning~\cite{kim2021cirn} or dynamic topology~\cite{raychaudhuri2022multi}.

\textbf{Interactive Multi-Objective Networks.}
A classical strategy towards building interactive networks for multiple objectives is based on weight/network interpolation~\cite{wang2018esrgan,wang2019dni,cai2021mod}. The main idea consists in training a network with the same topology multiple times using different losses. After training, the authors propose to produce all intermediate behaviors by interpolating corresponding network weights using a convex linear combination driven by a set of interactive interpolation parameters. This approach allows to use an arbitrary number of objectives and has negligible computational cost, however the results are often prone to artifacts as the interpolated weights are not explicitly supervised. Other approaches use feature modulation~\cite{shoshan2019dynet,wang2019cfsnet} to control the network behavior. These methods propose to use two separate branches connected at each layer via local residual connections. Each branch is trained in a separate stage against a different objective, while keeping the other one frozen. Although effective, this approach has several drawbacks: it is limited to two objectives, the intermediate behaviors are again not optimized, and the complexity is effectively doubled.

Compared to all existing works, our strategy is more flexible and more robust, as we can handle an arbitrary number of objectives and we explicitly optimize all their combinations. On top of that, our strategy is also straightforward to train, agnostic to the model architecture, and has negligible computational overhead at inference time. 

\section{Method}
\label{sec:method}

In this section, we formally introduce our framework to enable tunable network behaviour. As a starting point for our discussion, we recall the general form of traditional and dynamic convolutions (\cref{sec:method:background}). Next, we provide a formal definition for our tunable convolutions and discuss how to train these layers using the proposed parametric multi-loss optimization (\cref{sec:method:tune_net}).

\subsection{Background}
\label{sec:method:background}

\textbf{Traditional Convolutions.}
\label{sec:method:traditional}
Let us define the basic form of a traditional convolutional layer as 
\begin{equation}
    \boldsymbol{y} = f_c(\boldsymbol{x}) = \boldsymbol{x} \circledast \boldsymbol{k} + \boldsymbol{b},
    \label{eq:trad_conv}
\end{equation}
where $\circledast$ is a convolution with kernel $\boldsymbol{k} \in \mathbb{R}^{k \times k \times c \times d}$ and bias $\boldsymbol{b} \in \mathbb{R}^{d}$ which transforms an input
$\boldsymbol{x} \in \mathbb{R}^{h \times w \times c}$ into an output $\boldsymbol{y} \in \mathbb{R}^{h \times w \times d}$, being $h \times w$ the spatial resolution, $k$ the spatial kernel support, $c$ the input channels, and $d$ the output channels.

\textbf{Dynamic Convolutions.}
\label{sec:method:dynamic}
A dynamic convolution~\cite{chen2020dynamic,zhang2020dynet} 
\begin{equation}
    \boldsymbol{y} = f_d(\boldsymbol{x}) = \boldsymbol{x} \circledast \hat{\boldsymbol{k}}_{\boldsymbol{x}} + \hat{\boldsymbol{b}}_{\boldsymbol{x}},
    \label{eq:dyn_conv}
\end{equation}
is parametrized by a dynamic kernel and bias generated as
\begin{equation}
    \hat{\boldsymbol{k}}_{\boldsymbol{x}} = \sum_{i=1}^p \alpha_i \boldsymbol{k}_i \,\,\,\,\,\,\,\, \text{and} \,\,\,\,\,\,\,\, \hat{\boldsymbol{b}}_{\boldsymbol{x}} = \sum_{i=1}^p \alpha_i \boldsymbol{b}_i,
    \label{eq:dyn_aggregation}
\end{equation}
by aggregating an underlying set of fixed $p$ kernels and biases $\{\boldsymbol{k}_i, \boldsymbol{b}_i\}_{i=1}^p$, using $p$ aggregation weights $\boldsymbol{\alpha} = \{\alpha_i\}_{i=1}^p$ dynamically generated from the input. Note that this input-dependency is highlighted in~\eqref{eq:dyn_conv} via the subscript $\boldsymbol{x}$. Formally the aggregation weights can be defined as
\begin{equation}
    \boldsymbol{\alpha} = \phi_d(\boldsymbol{x}),
    \label{eq:dyn_weights}
\end{equation}
being $\phi_d: \mathbb{R}^{h \times w \times c} \rightarrow \mathbb{R}^p$ a function mapping the input $\boldsymbol{x}$ to the $p$ aggregation weights. This function is typically implemented as a squeeze-and-excitation (SE) layer~\cite{hu2018squeeze}, \ie a global pooling operation followed by a number of fully connected layers and a final softmax activation to ensure convexity (\ie $\sum_{i=1}^p \alpha_i = 1$). While this layer is capable of dynamically adapting its response, it still lacks a mechanism to control its behaviour in a predictable fashion.

\subsection{Tunable Networks}
\label{sec:method:tune_net}

In this section we will discuss the two core components proposed in this work, namely a new parametric layer, called tunable convolution, and a parametric optimization strategy to enable the sought-after tunable behavior. A schematic of our framework can be found in Fig.~\ref{fig:diagram}.

\input{figures/2_diagram}

\textbf{Tunable Convolutions.}
The first building block of our framework consists of defining a special form of dynamic \textit{and} tunable convolution
\begin{equation}
    \boldsymbol{y} = f_t(\boldsymbol{x}, \boldsymbol{\omega}) = \boldsymbol{x} \circledast \hat{\boldsymbol{k}}_{\boldsymbol{\omega}} + \hat{\boldsymbol{b}}_{\boldsymbol{\omega}},
    \label{eq:tune_conv}
\end{equation}
which includes an additional input $\boldsymbol{\omega} = \{\omega_i\}_{i=1}^p$ consisting of $p$ of interactive parameters which are used to control the effect of $p$ different objectives. The kernels and biases are aggregated analogously to~\eqref{eq:dyn_aggregation}. However there is a key difference: here we propose to generate the aggregation weights not implicitly from the input $\boldsymbol{x}$, but rather explicitly from the interactive parameters as
\begin{equation}
    \boldsymbol{\alpha} = \phi_t(\boldsymbol{\omega}),
    \label{eq:tune_weights}
\end{equation}
where $\phi_t: \mathbb{R}^p \rightarrow \mathbb{R}^p$ is a function mapping the $p$ interactive parameters into the $p$ aggregation weights. Crucially, the kernel and bias of~\eqref{eq:tune_conv} now depends on the input parameters, as denoted by the subscript $\boldsymbol{\omega}$, thus highlighting the differences with respect to dynamic convolutions~\eqref{eq:dyn_conv}. Without loss of generality, $\boldsymbol{\omega}$ are all assumed to be in the range $[0,1]$ but are not necessarily convex (\ie $\sum_{i=1}^p \omega_i \geq 1$). Intuitively, the influence of the $i$-th objective is minimized by $\omega_i=0$ and maximized by $\omega_i=1$. There are multiple possibilities for the actual implementation of $\phi_t$, ranging from SE~\cite{hu2018squeeze} to a simple identity function. In our experiments we use a learnable affine transformation (\ie an MLP) without any activation, as we empirically found that this leads to better solutions in all our experiments, and also has a minimal impact on computational complexity. From a practical point of view, the proposed tunable convolutions are agnostic to input dimensions or convolution hyper-parameters. Thus, as we will show in our experiments, tunable variants of strided, transposed, point-wise, group-wise convolutions, or even attention layers, can be easily implemented.

\input{figures/3_runtime}

There are several advantages that come with the proposed strategy: first, it increases the representation power of the model~\cite{chen2020dynamic,luo2021functional}; second, it generates the aggregation weights in a predictable and interpretable fashion from the external parameters, as opposed to~\eqref{eq:dyn_weights} which generates the non-interactive weights from the input; third, it is computationally efficient as the only additional cost over a traditional convolution consists in computing~\eqref{eq:tune_weights} and the corresponding kernel aggregation. Fig.~\ref{fig:runtime} shows the average increase in runtime\footnote{Measured by averaging $5000$ runs on a NVIDIA V100 GPU.} required to process a tunable convolution compared to a traditional convolution with equivalent kernel size and number of channels. As may be observed from the figure the overhead is less than $20\%$, \ie few fractions of a millisecond, and in the most common cases of $p \leq 4$ it is even lower than $5\%$.

\textbf{Parametric Multi-Loss.}
Let us outline how to instill tunable behavior in~\eqref{eq:tune_weights} via the interactive parameters $\boldsymbol{\omega}$. The tunable convolution needs to be informed of the meaning of each parameter and supervised accordingly as each parameter is changed. We achieve this by explicitly linking these parameters to different behaviors through the use of a parametric multi-loss function composed of $p$ different objectives $\mathcal{L}_i$. Specifically, the \emph{same} parameters $\boldsymbol{\omega}$ used in~\eqref{eq:tune_conv} to aggregate the tunable kernels and biases are also used to aggregate the $p$ objectives as
\begin{equation}
    \widehat{\mathcal{L}}_{\boldsymbol{\omega}} = \sum_{i=1}^{p} \omega_{i} \cdot \lambda_i \cdot \mathcal{L}_i,
    \label{eq:tune_loss}
\end{equation}
where $\lambda_i \geq 0$ is a fixed weight used to scale the relative contribution of the $i$-th objective to the overall loss. Note that each $\mathcal{L}_i$ can include multiple terms, so it is possible to encapsulate complex behaviors, such as a style transfer~\cite{johnson2016style} or GAN loss~\cite{wang2018esrgan}, into a single controllable objective.

Our approach generalizes a vanilla training strategy. Specifically, if we keep $\boldsymbol{\omega}$ fixed during training, then the corresponding loss~\eqref{eq:tune_loss} will also be fixed and thus no tunable capability will be explicitly induced by $\boldsymbol{\omega}$. Differently, we propose to also optimize all intermediate objectives by sampling at each training step a different, random, set of parameters, which we use to generate a random combination of objectives in~\eqref{eq:tune_loss}, and a corresponding combination of kernels in~\eqref{eq:tune_conv}. Thus, the network is encouraged to disentangle the different objectives into the different tunable kernels (and biases) and, as a result, at inference time we can control the relative importance of the different objectives by interacting with the corresponding parameters. The critical advantage over prior methods~\cite{wang2019dni,shoshan2019dynet,wang2019cfsnet,he2020cresmd} is that our strategy is not restricted to a fixed number of objectives, and also actively optimizes all their intermediate combinations. In this work we use random uniform sampling, \ie $\omega_i \sim \mathcal{U}(0,1)$, for both its simplicity and excellent empirical performance, however different distributions could be explored to bias the sampling towards specific objectives~\cite{he2020cresmd}.

\section{Experiments}
\label{sec:experiments}

In this section we report experimental results of our tunable convolutions evaluated on image denoising, deblurring, super-resolution, and style transfer. In all tasks, we measure the ability of our method to tune inference behavior by interacting with external parameters designed to control various characteristics of the image translation process. To showcase the efficacy of our method, we compare performance to recent controllable networks, namely DNI~\cite{wang2019dni}, CFSNet~\cite{wang2019cfsnet}, DyNet~\cite{shoshan2019dynet}, and CResMD~\cite{he2020cresmd}. Further, we use our tunable convolutions as drop-in replacements in the state-of-the-art SwinIR~\cite{liang2021swinir} and NAFNet~\cite{chen2022nafnet} networks and evaluate performance in standard benchmarks. A MindSpore~\cite{mindspore} reference implementation is available\footnote{\url{https://github.com/mindspore-lab/mindediting}}.

\input{figures/4_exp_dn_rec_noise}

\subsection{Image Restoration}

A general observation model for image restoration is
\begin{equation}
    z = D(y) + \eta,
    \label{eq:observation_model}
\end{equation}
where $z \in \mathbb{R}^{h \times w \times 3}$ is a degraded image with resolution $h{\times}w$ and three $RGB$ color channels, $D$ is a degradation operator (\eg downsampling or blurring) applied to the underlying (unknown) ground-truth image $y$, and $\eta$ is a random noise realization which models the stochastic nature of the image acquisition process. Under this lens, a restoration network aims to provide a faithful estimate $\hat{y}$ of the underlying ground-truth image $y$, given the corresponding degraded observation $z$.

\subsubsection{Denoising}
\label{sec:experiments:dn}
In this section we assess performance of our tunable convolutions used in different backbones applied to the classical problem of image denoising. Formally, if we refer to~\eqref{eq:observation_model}, $D$ is the identity, and $\eta$ is typically distributed as i.i.d.~Gaussian noise. An interesting application of tunable models looks into modulating the denoising strength to balance the amount of noise removal and detail preservation in the predicted image; two objectives that, because of the inherent imperfection of any denoising process, are often in conflict with one another. Formally, we use a multi-loss $\widehat{\mathcal{L}}_{\text{rn}} = \omega_1 \cdot \mathcal{L}_{\text{rec}} + \omega_2 \cdot \mathcal{L}_{\text{noise}}$ which includes two terms: the first is a standard reconstruction objective measuring distortion with an $L_1$ distance, and the second is a novel noise preservation objective defined as
\begin{equation}
    \mathcal{L}_{\text{noise}} = || \hat{y} - y_{\eta} ||_1,
    \label{eq:noise_loss}
\end{equation}
where $y_{\eta} = y + \omega_2 \cdot \nu \cdot (z - y)$ is the target image containing an amount of residual noise proportional to the parameter $\omega_2$ and a fixed pre-defined scalar $0 \leq \nu \leq 1$. Note that we set $\nu=0.9$, so that even with maximum noise preservation $\omega_2=1$ we avoid convergence to a trivial solution (\ie target image equal to the noisy input), and instead require $90\%$ of the residual noise to be preserved. Let us recall that different combinations of $(\omega_1, \omega_2)$ promote different inference behaviors, \eg $(0.00, 1.00)$ maximizes noise preservation and $(1.00, 0.00)$ maximizes fidelity. Finally, in order to objectively measure the tunable ability of the compared methods, we use PSNR against the ground-truth $y$ and $\text{PSNR}_\eta$ against the target image $y_{\eta}$ as clear performance indicators for $\mathcal{L}_{\text{rec}}$ and $\mathcal{L}_{\text{noise}}$, respectively.

\textbf{Tunable Networks.}
Here we compare against the state-of-the-art in controllable networks, namely DNI~\cite{wang2019dni}, DyNet~\cite{shoshan2019dynet}, and CFSNet~\cite{wang2019cfsnet} applied to synthetic image denoising. For fair comparison, we use the same ResNet~\cite{he2016resnet} backbone as in~\cite{shoshan2019dynet,wang2019cfsnet,he2020cresmd} which includes a long residual connection before the output layer. Specifically, for DNI and our tunable network, we stack $16$ residual blocks (Conv2d-ReLU-Conv2d-Skip) with $64$ channels, whereas for DyNet and CFSNet we use $8$ blocks for the main branch and $8$ blocks for the tuning branch in order to maintain the same overall complexity. All methods are trained for $500,000$ iterations using Adam optimizer~\cite{kingma2014adam} with batch size $16$ and fixed learning rate $1e{-4}$. For training we use patches of size $64 \times 64$, randomly extracted from the DIV2K~\cite{agustsson2017ntire} dataset, to which we add Gaussian noise $\eta \sim \mathcal{N}(0, \sigma^2)$ with standard deviation $\sigma \in [5, 30]$ as in~\eqref{eq:observation_model}.

\input{tables/1_exp_dn_rec_noise}

In Table~\ref{tbl:dn:strength} we report performance averaged over noise levels $\sigma \in [5, 15, 30]$. It may be observed that the proposed tunable model outperforms the state of the art in almost all cases. Further, our method has almost identical accuracy to a network purely trained for fidelity, \ie DNI with weights $(1.00, 0.00)$, whereas the others typically show a $-0.2$dB drop in PSNR. In general, DNI and DyNet do not generalize well in the intermediate cases, whereas CFSNet has competitive performance, especially in regimes where $\mathcal{L}_{\text{noise}}$ dominates. However the visual comparisons against CFSNet in Fig.~\ref{fig:dn:strength}, show that our method is characterized by a smoother and more consistent transition between objectives over the parameter range, and when denoising is maximal, our prediction also contains fewer artifacts.

\textbf{Traditional Networks.}
In this section, we build tunable variants of traditional (fixed) state-of-the-art networks SwinIR~\cite{liang2021swinir} and NAFNet~\cite{chen2022nafnet}. Note that these models contain a plethora of different layers such as MLPs, strided and depth-wise convolutions, as well as spatial, channel, and window attention, and all of these can be easily replaced by our tunable variants. We construct and train our tunable SwinIR and NAFNet models following the setup outlined in the original papers, which we also summarize in the supplementary materials.

\input{tables/2_exp_dn_rec_noise_sota}

In Table~\ref{tbl:dn:strength_swinir}, we report PSNR of our tunable SwinIR for color image denoising at noise levels $\sigma \in [15, 25, 50]$ against state-of-the-art IPT~\cite{chen2021ipt}, DRUNet~\cite{zhang2022drunet}, and fixed SwinIR. In Table~\ref{tbl:dn:strength_nafnet}, we report PSNR and SSIM~\cite{wang2004ssim} of our tunable NAFNet against the fixed NAFNet for real raw image denoising on SIDD~\cite{abdelrahman2018sidd}. In both tables, we show results for only two combinations of tuning parameters, one maximizing data fidelity $(0, 1)$ and the other maximizing noise preservation $(1, 0)$. Results largely concur with those in Table~\ref{tbl:dn:strength}, and show that our tunable networks have similar, and often even better, performance when compared to the corresponding fixed baselines trained purely for data fidelity.

\input{figures/5_exp_dn_db}

\subsubsection{Joint Denoising \& Deblurring}
\label{sec:experiments:jdb}
In this experiment we assess the ability of the proposed method to enable interactive image restoration across multiple degradations, specifically we consider joint denoising and deblurring. This is a task explored in CResMD~\cite{he2020cresmd} and as such we will use it here as baseline comparison. 

For training, we add synthetic Gaussian noise with standard deviation $\sigma \in [5, 30]$ as explained in the previous section, and, following~\cite{he2020cresmd}, we also add Gaussian blur using a $21{\times}21$ kernel with standard deviation in $\rho \in [0,4]$. Recalling~\eqref{eq:observation_model}, $D$ represents the blurring operation which is applied to the ground-truth $y$ before adding the noise $\eta$. We define a multi-loss to simultaneously control the amount of denoising and deblurring as $\widehat{\mathcal{L}}_{\text{nb}} = \omega_1 \cdot \mathcal{L}_{\text{noise}} + \omega_2 \cdot \mathcal{L}_{\text{blur}}$, which includes the same noise preservation objective of~\eqref{eq:noise_loss} plus a deblurring and sharpening objective $\mathcal{L}_{\text{blur}} = || \hat{y} - \tilde{y}_{\eta} ||_1$ where
\begin{equation*}
    \tilde{y}_{\eta} = y_{\eta} + \omega_2 \cdot \gamma \cdot (y_{\eta} - g \circledast y_{\eta})
\end{equation*}
is obtained via an ``unsharp mask filter''~\cite{polesel2000unsharp} using a Gaussian kernel $g$ with support $9 \times 9$ and standard deviation $2.5$ to extract the high-pass information from $y_{\eta}$ which is then scaled and added back to the same image to enhance edges and contrast. We introduce a fixed scalar $\gamma = 8$ to define the maximum amount of sharpening that can be applied to the image. Note that when $\gamma = 0$, the objective $\mathcal{L}_{\text{blur}}$ will be equivalent to $\mathcal{L}_{\text{noise}}$, as only deblurring but no sharpening will be applied to the input image.

\input{tables/3_exp_dn_db}

In this experiment, the tuning parameters $(\omega_1, \omega_2)$ are non-convex, and individually control the amount of noise and blur in the prediction. As can be seen from Fig.~\ref{fig:dn:deblur}, in our method these parameters explicitly control the influence of the noise and blur objectives; differently, in CResMD the parameters represent true degradation in the input, which are modified at inference time to change the restoration behaviour\footnote{For instance, assuming true noise level in the input is $\sigma$, using $\beta\cdot\sigma$ for some $\beta < 1$ would implicitly reduce the denoising strength.}.
%; for instance, assuming an image with $\sigma=15$ and blur size $\rho = 2$, if we aim to remove the correct amount of noise and blur we should use parameters $(\omega_1, \omega_2) = (0.5, 0.5)$ for CResMD and $(1.0, 0.0)$ for our method.
Thus, for the same values of parameters, the two methods generate outputs with different characteristics. In Table~\ref{tbl:dn:deblur} we report performance in terms of PNSR, LPIPS~\cite{zhang2018unreasonable}, and NIQE~\cite{mittal2013niqe} averaged over all levels of noise $[5, 15, 30]$ and blur $[0, 1, 2, 3, 4]$. These results clearly show that our tunable strategy outperforms CResMD almost everywhere, and often by a significant margin. Furthermore, as also noted in~\cite{kim2021cirn}, CResMD generates severe artifacts whenever asked to process combinations of input image and tuning parameters outside its training distribution (refer to the supplementary materials for some examples).

\subsubsection{Super-Resolution}
\label{sec:experiments:sr}

For this experiment we focus on the scenario of $\times 4$ super-resolution, and we consider the degradation operator $D$ in~\eqref{eq:observation_model} to be bicubic downsampling, and $\eta = 0$ (\ie no noise is added). We use the same setup as in the tunable denoising experiments, with the difference that this time for training we use patch size $48 \times 48$ (thus ground-truth patches are $192 \times 192$), and the backbone ResNet architecture includes a final pixel shuffling upsampling~\cite{shi2016real,lim2017edsr}.

\input{tables/4_exp_sr_rec_gan}
\input{figures/6_exp_sr_rec_gan}

Super-resolution is an ideal task to test the ability of our tunable model to balance the perception-distortion trade-off~\cite{blau2018perception}. We design a multi-loss $\widehat{\mathcal{L}}_{\text{rg}} = \omega_1 \cdot \mathcal{L}_{\text{rec}} + \omega_2 \cdot \mathcal{L}_{\text{gan}}$ which includes a reconstruction objective $\mathcal{L}_{\text{rec}}$ to maximize accuracy, and an adversarial objective $\mathcal{L}_{\text{gan}}$ to maximize perceptual quality. Following~\cite{wang2018esrgan}, the latter is defined as
$
    \mathcal{L}_{\text{gan}} = 0.01 \cdot \mathcal{L}_{\text{rec}} + \mathcal{L}_{\text{vgg}} + 0.005 \cdot \mathcal{L}_{\text{adv}}
$,
where $\mathcal{L}_{\text{vgg}}$ is a perceptual loss measuring $L_1$ distance of VGG-19~\cite{simonyan2015vgg} features obtained at the ``conv5\_4'' layer~\cite{johnson2016style,ledig2017loss}, and $\mathcal{L}_{adv}$ is a relativistic adversarial loss~\cite{martineau2019relgan} on the generator and discriminator. We use a discriminator similar to~\cite{wang2018esrgan}, with the difference that we use $4$ scales and we add a pooling operator before the final fully-connected layer in order for the classifier to work with arbitrary input resolutions.

\input{tables/5_exp_sr_rec_gan_sota}

\textbf{Tunable Networks.}
Objective results reported in Table~\ref{tbl:sr:gan} show that the proposed tunable strategy outperforms the compared methods almost everywhere in both fidelity and perceptual measures. Let us note that DyNet in this application exhibits better behavior and it is in general superior to CSFNet, whereas DNI still fails to correctly produce results corresponding to intermediate behaviors. Interestingly, our method is competitive, and even outperforms, a network solely trained with a reconstruction objective in terms of PSNR, \ie DNI with weights $(1.00, 0.00)$. In Fig.~\ref{fig:sr:gan} we show a visual example which demonstrate the smooth transition between the two objectives, and superior performance over DyNet in terms of quality of details and robustness to artifacts across the full parameter range.

\textbf{Traditional Networks.}
As in our denoising experiments, we also show comparison against the state of the art in classical $\times 4$ image super-resolution~\cite{dai2019san,mei2021nlsa,liang2021swinir}. In particular, we use SwinIR as backbone and we build a tunable version to optimize the same multi-loss $\widehat{\mathcal{L}}_{\text{rg}}$ described above; to ensure fair comparison, we use the training protocol outlined in~\cite{liang2021swinir} for input patch size $48 \times 48$. Results shown in Table~\ref{tbl:sr:gan_sota} confirms that, even in this arguably more challenging application, our tunable SwinIR achieves comparable results to the fixed baselines.

\subsection{Style Transfer}
In this section, in order to test a different image-to-image translation task as well as the the ability of our method to deal with more than two objectives, we consider style transfer~\cite{gatys2015style,johnson2016style}. Similarly to~\cite{johnson2016style}, we use a UNet~\cite{ronneberger2015unet} without skip connections consisting of two scales and $6$ residual blocks (Conv2d-ReLU-Conv2d-Skip-ReLU) to process latent features. Downsampling and upsampling is performed as a strided convolution and nearest neighbor interpolation, respectively. The number of channels is $64$ and then doubled at every scale. The first and last convolutions have kernel size $9 \times 9$. Instance normalization~\cite{ulyanov2016instance} and ReLU is applied after every convolution expect the last one. We train for $40,000$ iterations with weight decay $1e{-5}$, learning rate $1e{-4}$, batch size $4$, and patch size $384 \times 384$.

\input{figures/7_exp_style_mc_el_ky}

We define the style transfer objective similarly to~\cite{johnson2016style}: 
% $
%     \mathcal{L}_{\text{style}} = 0.03 \cdot \mathcal{L}_{\text{gram}} + 300 \cdot \mathcal{L}_{\text{vgg}} + 0.0001  \cdot \mathcal{L}_{\text{tv}} + \mathcal{L}_{\text{uv}}
% $, 
$
    \mathcal{L}_{\text{style}} = \lambda_{\text{gram}} \cdot \mathcal{L}_{\text{gram}} + \lambda_{\text{vgg}} \cdot \mathcal{L}_{\text{vgg}} + \lambda_{\text{tv}}  \cdot \mathcal{L}_{\text{tv}} + \mathcal{L}_{\text{uv}}
$,
where $\mathcal{L}_{\text{gram}}$ is the squared Frobenius norm of the difference between the Gram matrices of the predicted and target style features extracted from the ``relu3\_3'' layer of a pretrained VGG-19~\cite{simonyan2015vgg}, $\mathcal{L}_{\text{vgg}}$ is the $L_2$ distance between the predicted and target features extracted from the ``relu3\_3'' layer, $\mathcal{L}_{\text{tv}}$ is a Total Variation regularization term used to promote smoothness~\cite{aly2005tv}, and $\mathcal{L}_{\text{uv}}$ is the $L_2$ distance between the YUV chrominance channels used to preserve the original colors. We use different $\lambda$ weights for different styles; detailed settings can be found in the supplementary materials.

Fig.~\ref{fig:style} demonstrates that our method is capable to smoothly transition across three different styles, \ie \textit{Mosaic}, \textit{Edtaonisl}, and \textit{Kandinsky}. In contrast to existing approaches, such as DyNet~\cite{shoshan2019dynet}, our method is (the only one) able to reliably optimize more than two style objectives, including all their intermediate combinations.

\section{Discussion}
\label{sec:discussion}

In this section, we shed some light onto the inner workings of our method by analyzing the relationship between the kernels in our tunable convolution and the corresponding objectives in the multi-loss. As illustrative examples, we take the denoising and super-resolution experiments of Table~\ref{tbl:dn:strength} and Table~\ref{tbl:sr:gan}. In both cases we use the same ResNet backbone with $16$ residual blocks and $64$ channels. First, we tune these networks using five parameter combinations uniformly sampled with $0.25$ step; then we extract the \textit{tuned} kernel tensors from (both convolutions in) each residual block; and finally, we reduce the dimensionality via PCA followed by t-SNE~\cite{hinton2003sne,maaten2008tsne} to embed the tuned kernels into 2D points. In Fig.~\ref{fig:tsne:resblocks} we visualize scatter plots of these points for both denoising and super-resolution networks. Evidently, each block is well-separated into a different cluster, and, more interestingly, the tuned kernels span quasi-linear manifolds in the region identified by the tunable parameters with various lengths and orientations. This clearly demonstrates that the tuned kernels smoothly transition from one pure objective $(0.00,1.00)$ to the other $(1.00,0.00)$. The same linear transition can be also observed in Fig.~\ref{fig:pca:ker}, where we depict the first principal component of the input and output RGB kernels extracted from the first and last tunable convolution layer after tuning with $11$ parameters uniformly sampled with step $0.1$.

\input{figures/8_tsne_blocks}
\input{figures/9_kernels}

\section{Conclusions}
\label{sec:conclusions}

We have presented tunable convolutions: a novel dynamic layer enabling change of neural network inference behaviour via a set of interactive parameters. We associate each parameter with a desired behavior, or objective, in a multi-loss.
During training, these parameters are randomly sampled, and all possible combinations of objectives are explicitly optimized. 
During inference, the different objectives are disentangled into the corresponding parameters, which thus offer a clear interpretation as to which behaviour they should promote or inhibit. In comparison with existing solutions our strategy achieves better performance, is not limited to a fixed number of objectives, explicitly optimizes for all possible linear combinations of objectives, and has negligible computational cost. Further, we have shown that our tunable convolutions can be used as a drop-in replacement in existing state-of-the-art architecture enabling tunable behavior at almost no loss in baseline performance.

%%%%%%%%% REFERENCES
{\small
\bibliographystyle{ieee_fullname}
\bibliography{egbib}
}

\end{document}

%% file: figures/1_teaser.tex
\begin{figure}[t]
    
    % trim=left bottom right top
    % \newcommand{\tsrincludegraphics}[1]{\includegraphics[trim=0px 0px 70px 0px, clip, width=0.19\columnwidth]{images/1_teaser/#1}}
    \newcommand{\tsrincludegraphics}[1]{\includegraphics[width=0.19\columnwidth]{images/1_teaser/camera_ready/#1}}
    
    \setlength{\tabcolsep}{1pt}
    \scriptsize
    \centering
    \begin{tabular}{ccccc}
        \multicolumn{2}{l}{$\leftarrow$ Perceptual Quality} & & \multicolumn{2}{r}{Fidelity $\rightarrow$} \\[0.15cm]
        \tsrincludegraphics{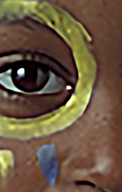} &
        \tsrincludegraphics{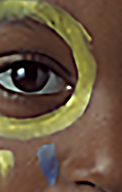} &
        \tsrincludegraphics{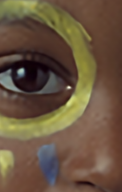} &
        \tsrincludegraphics{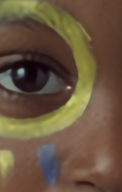} &
        \tsrincludegraphics{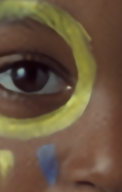} 
    \end{tabular}
    \caption{We propose a framework to build a single neural network that can be \emph{tuned} at inference without retraining by interacting with controllable parameters, \eg to balance the perception-distortion tradeoff in image restoration tasks.}
    \label{fig:teaser}
\end{figure}

%% file: figures/2_diagram.tex
\begin{figure}[t]
    \centering
    \includegraphics[trim=0px 125px 205px 0px,clip,width=0.7\columnwidth]{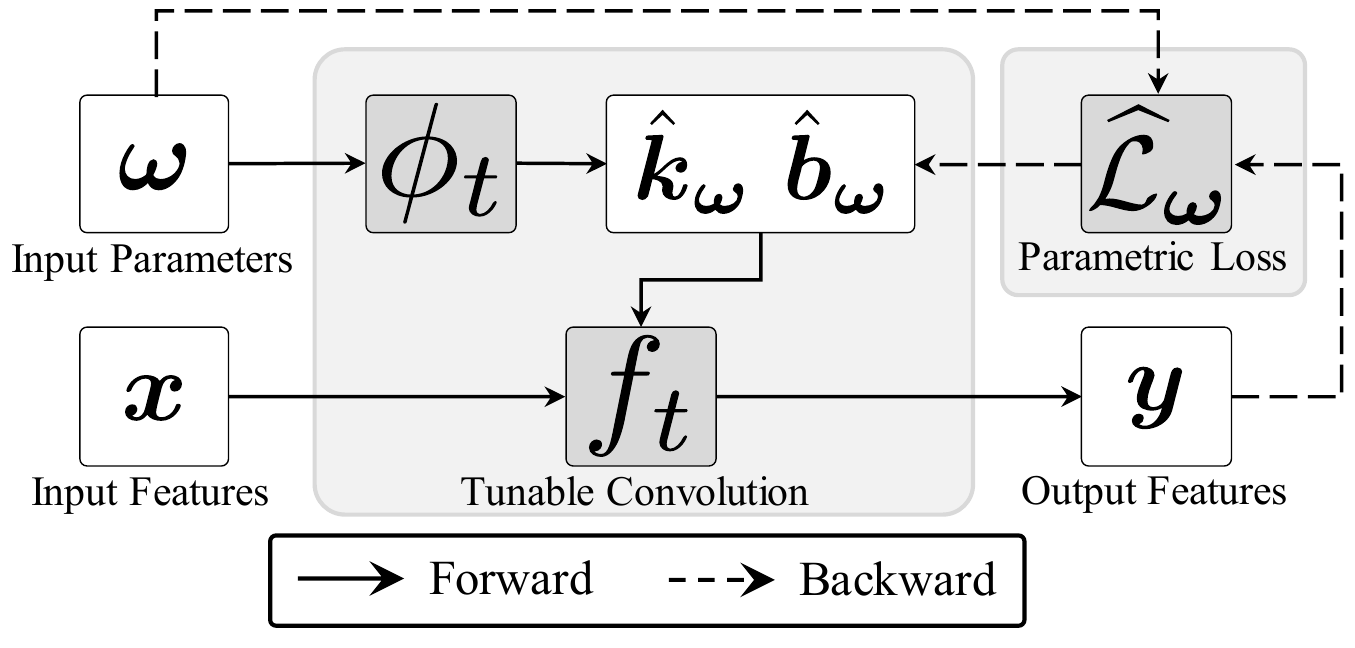}
    \caption{
    Illustration of the proposed framework to enable tunable networks, consisting of a tunable convolution taking as input a feature $\boldsymbol{x}$ and some external parameters $\boldsymbol{\omega}$ which also parametrize the loss used to update the tunable kernels and biases. See Sec.~\ref{sec:method:tune_net} for more details.
    }
    \label{fig:diagram}
\end{figure}

%% file: figures/3_runtime.tex
\begin{figure}[t]
    \centering
    \includegraphics[width=0.8\columnwidth]{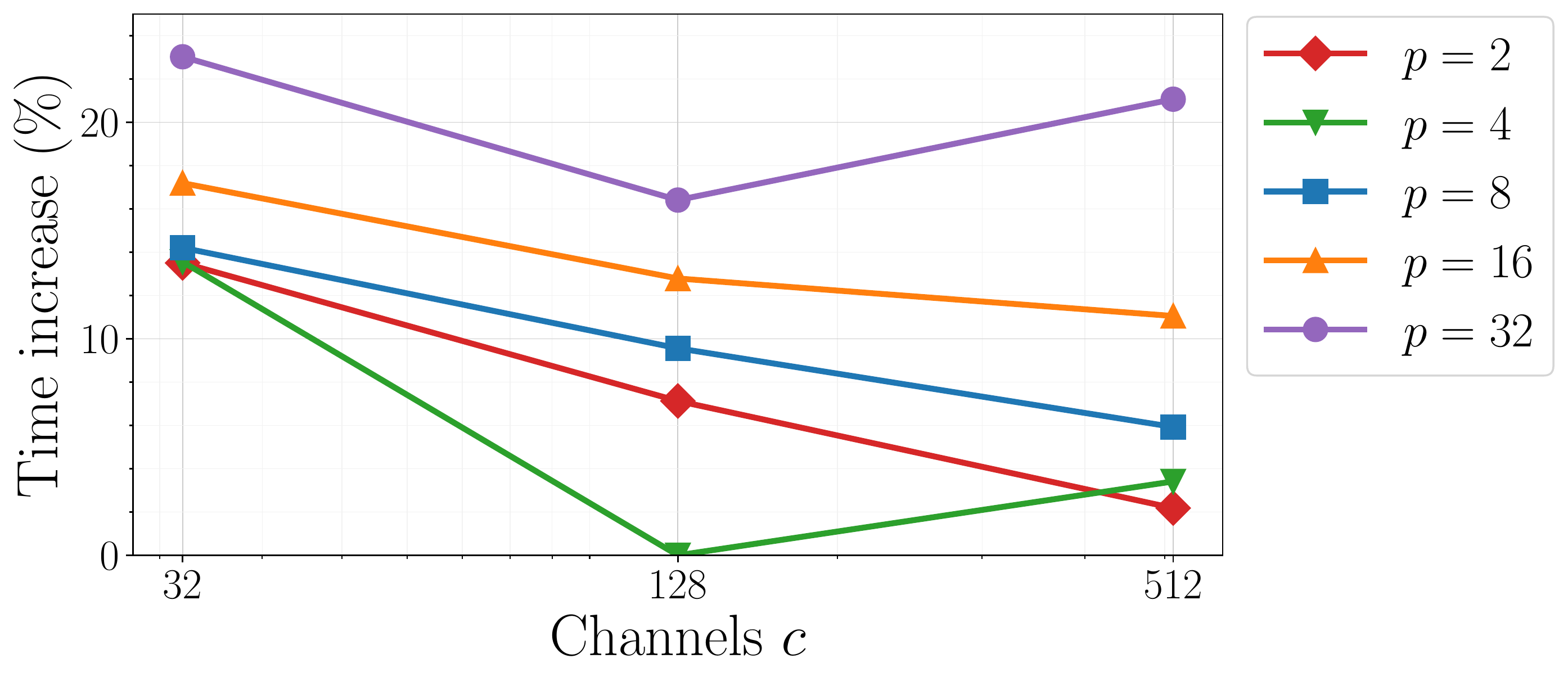}
    \caption{
    Average runtime increase (in percentage) between a traditional convolution with kernel size $k \in [3, 5, 7]$ and a tunable convolution with $p$ parameters while processing input of size $128 \times 128$ with varying number of channels $c$.
    }
    \label{fig:runtime}
\end{figure}

%% file: figures/4_exp_dn_rec_noise.tex
\begin{figure*}[!ht]

    % trim=left bottom right top
    % \newcommand{\dnsincludegraphics}[1]{\includegraphics[trim=0px 110px 0px 10px, clip, width=0.13\textwidth]{images/4_dn_rec_noise/#1}}
    \newcommand{\dnsincludegraphics}[1]{\includegraphics[width=0.13\textwidth]{images/4_dn_rec_noise/camera_ready/#1}}
    
    \centering
    \scriptsize
    \setlength{\tabcolsep}{2pt}
    \begin{tabular}{cc c ccccc c}
         & & \hspace{0.2cm} & & \multicolumn{2}{l}{$\leftarrow$ Less Denoising} & $(\mathcal{L}_{\text{rec}}, \mathcal{L}_{\text{noise}})$ & \multicolumn{2}{r}{More Denoising $\rightarrow$} \\[0.1cm]
         %\cline{2-6}
         & & & & $(0.00, 1.00)$ & $(0.25, 0.75)$ & $(0.50, 0.50)$ & $(0.75, 0.25)$ & $(1.00, 0.00)$ \\[0.1cm]
        
        \multirow{1}{*}[1.15cm]{\rotatebox[origin=c]{90}{{\scriptsize Ground-truth}}} &
        \dnsincludegraphics{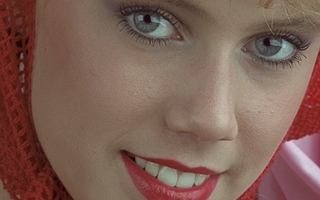} & &
        \multirow{1}{*}[1.10cm]{\rotatebox[origin=c]{90}{{\scriptsize CFSNet~\cite{wang2019cfsnet}}}} &
        \dnsincludegraphics{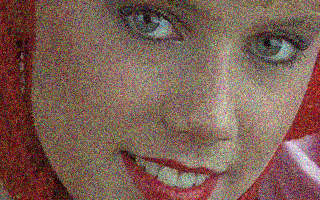} &
        \dnsincludegraphics{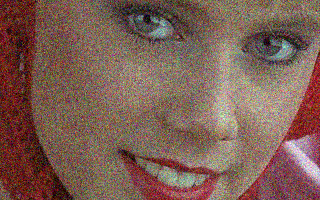} &
        \dnsincludegraphics{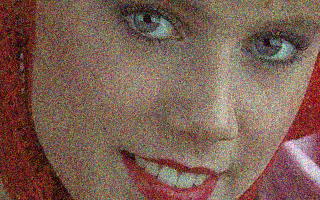} &
        \dnsincludegraphics{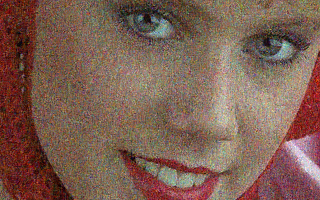} &
        \dnsincludegraphics{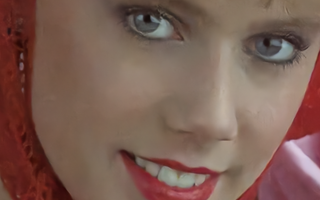} 
        
        \\
        
        \multirow{1}{*}[0.75cm]{\rotatebox[origin=c]{90}{{\scriptsize Input}}} &
        \dnsincludegraphics{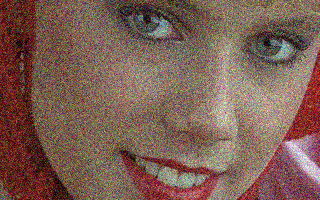} &  &
        \multirow{1}{*}[0.75cm]{\rotatebox[origin=c]{90}{{\scriptsize Ours}}} &
        \dnsincludegraphics{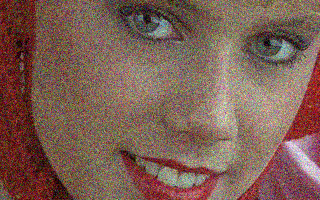} &
        \dnsincludegraphics{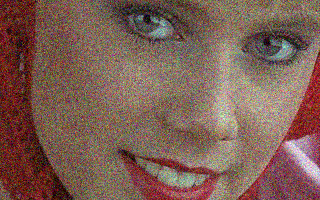} &
        \dnsincludegraphics{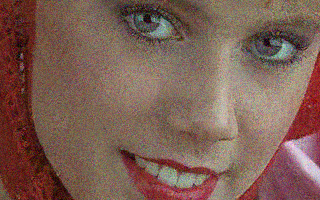} &
        \dnsincludegraphics{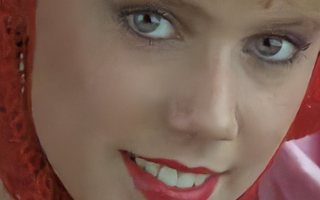} &
        \dnsincludegraphics{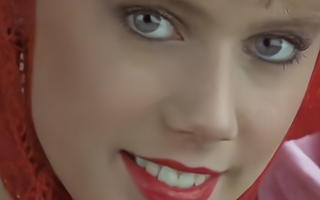}
    \end{tabular}
    \vspace{-0.25cm}
    \caption{Tuning denoising strength on image 04 from the Kodak dataset~\cite{kodak} corrupted by Gaussian noise with standard deviation $\sigma = 30$
    }
    \label{fig:dn:strength}
    \vspace{-0.25cm}
\end{figure*}

%% file: tables/1_exp_dn_rec_noise.tex
\begin{table}[t]
\footnotesize
\setlength{\tabcolsep}{2pt}

\centering
\begin{adjustbox}{scale=0.84}
    \begin{tabular}{x{0.4cm}ccccc c ccccc c}
        & \multicolumn{5}{c}{{\scriptsize Kodak~\cite{kodak}}} & & \multicolumn{5}{c}{{\scriptsize CBSD68~\cite{martin2001bsd68}}} \\
        \cmidrule(r){1-1} \cmidrule{2-6} \cmidrule{8-12} \cmidrule(l){13-13}
        {\scriptsize $\omega_1$ } 
        & {\scriptsize $0.00$ } 
        & {\scriptsize $0.25$ } 
        & {\scriptsize $0.50$ } 
        & {\scriptsize $0.75$ } 
        & {\scriptsize $1.00$ } &  
        & {\scriptsize $0.00$ } 
        & {\scriptsize $0.25$ } 
        & {\scriptsize $0.50$ } 
        & {\scriptsize $0.75$ } 
        & {\scriptsize $1.00$ } 
        & {\scriptsize $\mathcal{L}_{\text{rec}}$ } \\
        
        {\scriptsize $\omega_2$ } 
        & {\scriptsize $1.00$ } 
        & {\scriptsize $0.75$ } 
        & {\scriptsize $0.50$ } 
        & {\scriptsize $0.25$ } 
        & {\scriptsize $0.00$ } &  
        & {\scriptsize $1.00$ } 
        & {\scriptsize $0.75$ } 
        & {\scriptsize $0.50$ } 
        & {\scriptsize $0.25$ } 
        & {\scriptsize $0.00$ } 
        & {\scriptsize $\mathcal{L}_{\text{noise}}$ } \\
        
        \cmidrule(r){1-1} \cmidrule{2-6} \cmidrule{8-12} \cmidrule(l){13-13}
        
        \multirow{4}{*}{\rotatebox[origin=c]{90}{{\scriptsize PSNR}}}
         & \textbf{26.74} & 20.69 & 18.95 & 20.92 & \textbf{35.56} & 
         & \textbf{26.78} & 20.18 & 18.36 & 20.18 & \textbf{34.55} & {\scriptsize DNI~\cite{wang2019dni} } \\

         & 26.72 & 24.60 & 24.31 & 26.76 & 35.32 & 
         & 26.76 & 24.55 & 24.18 & 26.48 & 34.41 & {\scriptsize DyNet~\cite{shoshan2019dynet} } \\

         & 26.73 & 26.83 & 27.95 & 29.89 & 35.32 & 
         & 26.77 & 26.88 & 27.95 & 29.66 & 34.41 & {\scriptsize CFSNet~\cite{wang2019cfsnet} } \\
        
        \rowcolor{LightYellow} \cellcolor{White} 
         & \textbf{26.74} & \textbf{27.09} & \textbf{28.46} & \textbf{35.07} & 35.54 & \cellcolor{White} 
         & \textbf{26.78} & \textbf{27.12} & \textbf{28.12} & \textbf{34.10} & 34.53 & {\scriptsize Ours } \\
        
        \cmidrule(r){1-1} \cmidrule{2-6} \cmidrule{8-12} \cmidrule(l){13-13}
        
        \multirow{4}{*}{\rotatebox[origin=c]{90}{{\scriptsize $\hspace{0.05cm}\text{PSNR}_{\eta}$}}}
         & \textbf{52.21} & 21.74 & 19.12 & 21.00 & 35.32 & 
         & \textbf{51.55} & 21.16 & 18.53 & 20.27 & 34.41 & {\scriptsize DNI~\cite{wang2019dni} } \\

         & 50.32 & 28.69 & 26.46 & 28.11 & 35.26 & 
         & 49.89 & 28.39 & 26.19 & 27.75 & 33.61 & {\scriptsize DyNet~\cite{shoshan2019dynet} } \\

         & 51.83 & 38.49 & 34.45 & 32.88 & 35.32 & 
         & 51.33 & 38.45 & 34.33 & 32.59 & 34.41 & {\scriptsize CFSNet~\cite{wang2019cfsnet} } \\
        
        \rowcolor{LightYellow} \cellcolor{White} 
         & 51.97 & \textbf{39.04} & \textbf{35.56} & \textbf{36.19} & \textbf{35.54} & \cellcolor{White} 
         & 51.36 & \textbf{38.92} & \textbf{34.98} & \textbf{35.55} & \textbf{34.53} & {\scriptsize Ours } \\
        
        \cmidrule(r){1-1} \cmidrule{2-6} \cmidrule{8-12} \cmidrule(l){13-13}
        
    \end{tabular}
\end{adjustbox}

\caption{
%Tuning denoising from minimum $(\omega_1, \omega_2) = (0.00, 1.00)$ to maximum strength $(1.00, 0.00)$ to optimize for noise preservation and data fidelity, respectively.
Tuning denoising strength. Increasing parameter $\omega_1$ promotes data fidelity $\mathcal{L}_{\text{rec}}$, whereas increasing $\omega_2$ promotes noise preservation $\mathcal{L}_{\text{noise}}$. Our strategy is the best in almost all cases.
}
\label{tbl:dn:strength}
\end{table}

%% file: tables/2_exp_dn_rec_noise_sota.tex
\begin{table}[t]
\setlength{\tabcolsep}{2pt}

\begin{subtable}{0.54\columnwidth}
\begin{adjustbox}{scale=0.7,center}
    \centering
    \begin{tabular}{c cc ccc c}
        & {\small \hspace{0.025cm}IPT\hspace{0.025cm}} 
        & {\small DRUNet} 
        & \multicolumn{3}{c}{{\small SwinIR~\cite{liang2021swinir} } } \\
    
        {\normalsize $\sigma$ }
        & {\small \cite{chen2021ipt} }
        & {\small \cite{zhang2022drunet} }
        & {\scriptsize Fixed}
        & \cellcolor{LightYellow} {\scriptsize $(1, 0)$ }
        & \cellcolor{LightYellow} {\scriptsize $(0, 1)$ } \\
    
        \cmidrule{2-2} \cmidrule(lr){3-3} \cmidrule{4-6}
    
        % \textcolor{red}{}
        % \textcolor{blue}{}
        
        \multirow{1}{*}[0.025cm]{{\scriptsize 15 }}
        &  -   & 34.30 & \textbf{34.42} 
        & \cellcolor{LightYellow} 34.36 & \cellcolor{LightYellow} 54.33
        & \multirow{3}{*}[0.1cm]{\rotatebox[origin=c]{90}{{\scriptsize CBSD68~\cite{martin2001bsd68}}}} \\
         
        \multirow{1}{*}[0.025cm]{{\scriptsize 25 }}
        &  -   & 31.69 & \textbf{31.78} 
        & \cellcolor{LightYellow} 31.69 & \cellcolor{LightYellow} 51.67 & \\
         
        \multirow{1}{*}[0.025cm]{{\scriptsize 50 }}
        & 28.39 & 28.51 & \textbf{28.56} 
        & \cellcolor{LightYellow} 28.43 & \cellcolor{LightYellow} 48.42 & \\
        
        \cmidrule{2-2} \cmidrule(lr){3-3} \cmidrule{4-6}
    
        \multirow{1}{*}[0.025cm]{{\scriptsize 15 }}
        &  -   & 35.31 & 35.34
        & \cellcolor{LightYellow} \textbf{35.41} & \cellcolor{LightYellow} 55.41 
        & \multirow{3}{*}[0.025cm]{\rotatebox[origin=c]{90}{{\scriptsize Kodak~\cite{kodak}}}} \\
         
        \multirow{1}{*}[0.025cm]{{\scriptsize 25 }}
        &  -   & 32.89 & 32.89
        & \cellcolor{LightYellow} \textbf{32.94} & \cellcolor{LightYellow} 52.93 & \\
         
        \multirow{1}{*}[0.025cm]{{\scriptsize 50 }}
        & 29.64 & 29.86 & 29.79
        & \cellcolor{LightYellow} \textbf{29.86} & \cellcolor{LightYellow} 49.84 & \\

        \cmidrule{2-2} \cmidrule(lr){3-3} \cmidrule{4-6}
        
        & \multirow{1}{*}[0.15cm]{{\tiny PSNR}}
        & \multirow{1}{*}[0.15cm]{{\tiny PSNR}}
        & \multirow{1}{*}[0.15cm]{{\tiny PSNR}}
        & \multirow{1}{*}[0.15cm]{{\tiny PSNR}}
        & \multirow{1}{*}[0.15cm]{{\tiny PSNR$_\eta$}}
        & 
    
    \end{tabular}
\end{adjustbox}
\vspace{-0.2cm}
\caption{Color image denoising.}
\label{tbl:dn:strength_swinir}
\end{subtable}
\begin{subtable}{0.45\columnwidth}
\begin{adjustbox}{scale=0.7,center}
    \centering
    \begin{tabular}{l ccc c}
        & & & & \\[0.425cm]

        & \multicolumn{3}{c}{{\small NAFNet}~\cite{chen2022nafnet} } & \\
        
        % \cmidrule{3-5}
        
        & {\scriptsize Fixed}
        & \cellcolor{LightYellow} {\scriptsize $(1, 0)$ }
        & \cellcolor{LightYellow} {\scriptsize $(0, 1)$ }
        & \\

        \cmidrule(r){1-1} \cmidrule{2-4}
        
        \multirow{1}{*}[0.05cm]{{\tiny PSNR }}
        & \textbf{39.96} & \cellcolor{LightYellow} 39.90 & \cellcolor{LightYellow} 24.55 
        & \multirow{4}{*}[0.025cm]{\rotatebox[origin=c]{90}{{\scriptsize SIDD~\cite{abdelrahman2018sidd}}} } \\ 
        \multirow{1}{*}[0.05cm]{{\tiny PSNR$\eta$ }}
        & - & \cellcolor{LightYellow} 39.90 & \cellcolor{LightYellow} 59.90 & \\ 

        \cmidrule(r){1-1} \cmidrule{2-4}
        
        \multirow{1}{*}[0.05cm]{{\tiny SSIM }}
        & \textbf{0.960} & \cellcolor{LightYellow} \textbf{0.960} & \cellcolor{LightYellow} 0.523 & \\ 
        \multirow{1}{*}[0.05cm]{{\tiny SSIM$\eta$ }}
        & - & \cellcolor{LightYellow} 0.960 & \cellcolor{LightYellow} 0.999 & \\ 

        \cmidrule(r){1-1} \cmidrule{2-4} \\[0cm]
        
    \end{tabular}
\end{adjustbox}
\vspace{-0.2cm}
\caption{Real raw image denoising.}
\label{tbl:dn:strength_nafnet}
\end{subtable}
\vspace{-0.215cm}
\caption{
SwinIR and NAFNet trained for tunable denoising strength ($\mathcal{L}_{\text{rec}}$,~$\mathcal{L}_{\text{noise}}$). Tuning for data fidelity $(1, 0)$ results in performance comparable to the fixed baselines.
}
\label{tbl:dn:strength_sota}
\end{table}

%% file: figures/5_exp_dn_db.tex
\begin{figure}[!t]

    % trim=left bottom right top
    % \newcommand{\ddbincludegraphics}[1]{\includegraphics[trim=0px 0px 0px 70px, clip, width=0.22\columnwidth]{images/5_dn_db/#1}}
    \newcommand{\ddbincludegraphics}[1]{\includegraphics[width=0.22\columnwidth]{images/5_dn_db/camera_ready/#1}}
    
    \newcommand{\rspace}{-0.05cm}
    
    \centering
    \scriptsize
    \setlength{\tabcolsep}{1pt}
    \begin{tabular}{c c ccc c c}
        & \hspace{0.2cm} & 
        \multicolumn{1}{l}{{\tiny $\leftarrow$ Less Deblurring}} & 
        {\tiny $\mathcal{L}_{\text{blur}}$} & 
        \multicolumn{2}{r}{{\tiny More Deblurring $\rightarrow$}} & \\[0.075cm]
        
        % \cmidrule{3-5}

        \ddbincludegraphics{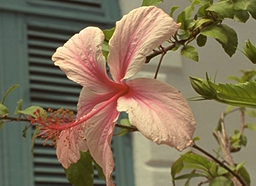} & &
        
        \ddbincludegraphics{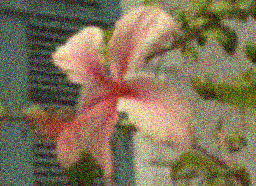} &
        \ddbincludegraphics{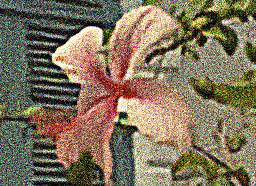} &
        \ddbincludegraphics{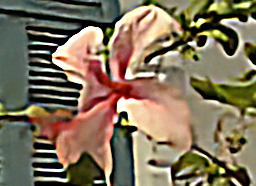} & &
        \multirow{1}{*}[1.1cm]{\rotatebox[origin=c]{270}{{\tiny $\leftarrow$ Less Denoising}}}
        
        \\[\rspace]
        
        Ground-truth & & 

        $(0.00,0.00)$ & $(0.00,0.50)$ & $(0.00,1.00)$ & &
        
        \\[0.05cm]
        
        \ddbincludegraphics{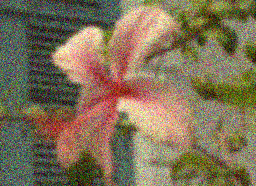} & &
        
        \ddbincludegraphics{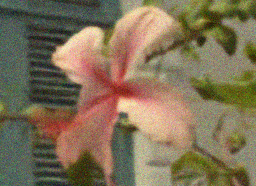} &
        \ddbincludegraphics{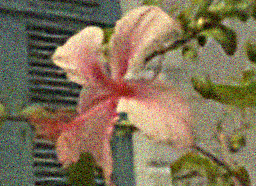} &
        \ddbincludegraphics{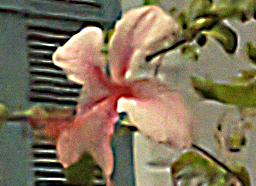} & &
        \multirow{1}{*}[0.725cm]{\rotatebox[origin=c]{270}{{\tiny $\mathcal{L}_{\text{noise}}$}}}
        
        \\[\rspace]
        
        Input & & 
        
        $(0.50,0.00)$ & $(0.50,0.50)$ & $(0.50,1.00)$ & &
        
        \\[0.05cm]
        
         & &
        
        \ddbincludegraphics{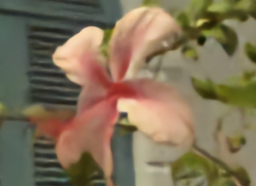} &
        \ddbincludegraphics{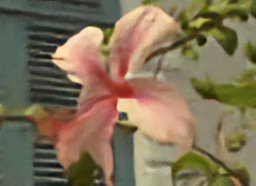} &
        \ddbincludegraphics{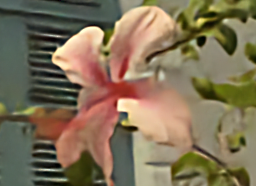} & &
        \multirow{1}{*}[1.2cm]{\rotatebox[origin=c]{270}{{\tiny More Denoising $\rightarrow$}}}
        
        \\[\rspace]
        
         & & 
        
        $(1.00,0.00)$ & $(1.00,0.50)$ & $(1.00,1.00)$ & \multicolumn{2}{c}{}
        
    \end{tabular}
    
    \vspace{-0.25cm}
    \caption{Image 07 from the Kodak dataset~\cite{kodak} corrupted by Gaussian noise $\sigma = 30$ and Gaussian blur with size $\rho=2$. Our method optimizes denoising and deblurring objectives for all combinations of parameters $(\omega_1, \omega_2)$.
    }
    \label{fig:dn:deblur}
    \vspace{-0.25cm}
\end{figure}

%% file: tables/3_exp_dn_db.tex
\begin{table}[t]
\footnotesize
\setlength{\tabcolsep}{3pt}

\centering
\begin{adjustbox}{scale=0.83}
    \begin{tabular}{x{0.5cm}cccc c cccc c}
        & \multicolumn{4}{c}{{\scriptsize Kodak~\cite{kodak}}} & & \multicolumn{4}{c}{{\scriptsize CBSD68~\cite{martin2001bsd68}}} \\
        \cmidrule(r){1-1} \cmidrule{2-5} \cmidrule{7-10} \cmidrule(l){11-11}
        {\scriptsize $\omega_1$ } 
        & {\scriptsize $0.00$ } 
        & {\scriptsize $0.00$ } 
        & {\scriptsize $1.00$ } 
        & {\scriptsize $1.00$ } &  
        & {\scriptsize $0.00$ }
        & {\scriptsize $0.00$ } 
        & {\scriptsize $1.00$ } 
        & {\scriptsize $1.00$ } 
        & {\scriptsize Noise $\eta$ } \\
        
        {\scriptsize $\omega_2$ } 
        & {\scriptsize $0.00$ } 
        & {\scriptsize $1.00$ } 
        & {\scriptsize $0.00$ } 
        & {\scriptsize $1.00$ } &  
        & {\scriptsize $0.00$ } 
        & {\scriptsize $1.00$ } 
        & {\scriptsize $0.00$ } 
        & {\scriptsize $1.00$ } 
        & {\scriptsize Blur $D$ } \\
        
        \cmidrule(r){1-1} \cmidrule{2-5} \cmidrule{7-10} \cmidrule(l){11-11}
        
        \multirow{2}{*}{\rotatebox[origin=c]{90}{{\scriptsize PSNR}}}
         & \textbf{23.68} & 11.75 & 25.92 & 20.65 & 
         & \textbf{23.14} & 11.84 & 24.84 & 19.97 & {\scriptsize CResMD~\cite{he2020cresmd} } \\
        
         & \cellcolor{LightYellow} 23.27 & \cellcolor{LightYellow} \textbf{18.15} & \cellcolor{LightYellow} \textbf{28.05} & \cellcolor{LightYellow} \textbf{24.84} & \cellcolor{White} 
         & \cellcolor{LightYellow} 22.17 & \cellcolor{LightYellow} \textbf{17.20} & \cellcolor{LightYellow} \textbf{27.13} & \cellcolor{LightYellow} \textbf{23.86} & \cellcolor{LightYellow} {\scriptsize Ours } \\
        
        \cmidrule(r){1-1} \cmidrule{2-5} \cmidrule{7-10} \cmidrule(l){11-11}

        \multirow{2}{*}{\rotatebox[origin=c]{90}{{\scriptsize LPIPS}}}
         & 0.190 & 0.580 & 0.127 & 0.218 & 
         & \textbf{0.195} & 0.586 & 0.144 & 0.248 & {\scriptsize CResMD~\cite{he2020cresmd} } \\
        
         & \cellcolor{LightYellow} \textbf{0.172} & \cellcolor{LightYellow} \textbf{0.381} & \cellcolor{LightYellow} \textbf{0.124} & \cellcolor{LightYellow} \textbf{0.205} & \cellcolor{White} 
         & \cellcolor{LightYellow} 0.203 & \cellcolor{LightYellow} \textbf{0.428} & \cellcolor{LightYellow} \textbf{0.136} & \cellcolor{LightYellow} \textbf{0.233} & \cellcolor{LightYellow} {\scriptsize Ours } \\
        
        \cmidrule(r){1-1} \cmidrule{2-5} \cmidrule{7-10} \cmidrule(l){11-11}

        \multirow{2}{*}{\rotatebox[origin=c]{90}{{\scriptsize NIQE}}}
         & \textbf{13.54} & 17.90 & 14.53 & \textbf{13.62} & 
         & \textbf{15.83} & 21.03 & 18.27 & \textbf{17.64} & {\scriptsize CResMD~\cite{he2020cresmd} } \\
        
         & \cellcolor{LightYellow} 16.18 & \cellcolor{LightYellow} \textbf{11.08} & \cellcolor{LightYellow} \textbf{13.98} & \cellcolor{LightYellow} 14.13 & \cellcolor{White} 
         & \cellcolor{LightYellow} 19.02 & \cellcolor{LightYellow} \textbf{14.31} & \cellcolor{LightYellow} \textbf{18.08} & \cellcolor{LightYellow} 18.87 & \cellcolor{LightYellow} {\scriptsize Ours } \\
        
        \cmidrule(r){1-1} \cmidrule{2-5} \cmidrule{7-10} \cmidrule(l){11-11}
        
    \end{tabular}
\end{adjustbox}
\vspace{-0.2cm}
\caption{
Tuning denoising and deblurring. In CResMD, the parameters $(\omega_1, \omega_2)$ are considered as the amount of noise and blur present in the input; in our methods they are used to explicitly con-\linebreak[4]trol the amount of noise and blur to remove.
}
\label{tbl:dn:deblur}
\end{table}

%% file: tables/4_exp_sr_rec_gan.tex
\begin{table}[t]
\footnotesize
\setlength{\tabcolsep}{2pt}

\centering
\begin{adjustbox}{scale=0.84}
    \begin{tabular}{x{0.4cm}ccccc c ccccc c}
        & \multicolumn{5}{c}{{\scriptsize Kodak~\cite{kodak}}} & & \multicolumn{5}{c}{{\scriptsize CBSD68~\cite{martin2001bsd68}}} \\
        \cmidrule(r){1-1} \cmidrule{2-6} \cmidrule{8-12} \cmidrule(l){13-13}
        {\scriptsize $\omega_1$ } 
        & {\scriptsize 0.00 } 
        & {\scriptsize 0.25 } 
        & {\scriptsize 0.50 } 
        & {\scriptsize 0.75 } 
        & {\scriptsize 1.00 } &  
        & {\scriptsize 0.00 } 
        & {\scriptsize 0.25 } 
        & {\scriptsize 0.50 } 
        & {\scriptsize 0.75 } 
        & {\scriptsize 1.00 } 
        & {\scriptsize $\mathcal{L}_{\text{rec}}$ } \\
        
        {\scriptsize $\omega_2$ } 
        & {\scriptsize $1.00$ } 
        & {\scriptsize $0.75$ } 
        & {\scriptsize $0.50$ } 
        & {\scriptsize $0.25$ } 
        & {\scriptsize $0.00$ } &  
        & {\scriptsize $1.00$ } 
        & {\scriptsize $0.75$ } 
        & {\scriptsize $0.50$ } 
        & {\scriptsize $0.25$ } 
        & {\scriptsize $0.00$ } 
        & {\scriptsize $\mathcal{L}_{\text{gan}}$ } \\
        
        \cmidrule(r){1-1} \cmidrule{2-6} \cmidrule{8-12} \cmidrule(l){13-13}
        
        \multirow{4}{*}{\rotatebox[origin=c]{90}{{\scriptsize PSNR}}}
         & 24.34 & 17.89 & 15.63 & 17.10 & 27.25 & 
         & 22.88 & 17.17 & 14.83 & 16.27 & \textbf{26.25} & {\scriptsize DNI~\cite{wang2019dni} } \\

         & 25.34 & 26.54 & 27.08 & 27.22 & 27.31 & 
         & 23.99 & 25.24 & 25.82 & 26.00 & 26.08 & {\scriptsize DyNet~\cite{shoshan2019dynet} } \\

         & \textbf{25.54} & 26.31 & 26.87 & 27.20 & 27.31 & 
         & \textbf{24.08} & 24.98 & 25.62 & 25.98 & 26.08 & {\scriptsize CFSNet~\cite{wang2019cfsnet} } \\
        
        \rowcolor{LightYellow} \cellcolor{White} 
         & 24.96 & \textbf{26.72} & \textbf{27.12} & \textbf{27.28} & \textbf{27.37} & \cellcolor{White} 
         & 23.61 & \textbf{25.42} & \textbf{25.82} & \textbf{26.05} & 26.10 & {\scriptsize Ours } \\
        
        \cmidrule(r){1-1} \cmidrule{2-6} \cmidrule{8-12} \cmidrule(l){13-13}
        
        \multirow{4}{*}{\rotatebox[origin=c]{90}{{\scriptsize LPIPS}}}
         & 0.113 & 0.114 & 0.114 & 0.114 & 0.106 & 
         & 0.130 & 0.141 & 0.141 & 0.141 & 0.132 & {\scriptsize DNI~\cite{wang2019dni} } \\

         & 0.103 & 0.096 & 0.102 & 0.105 & 0.106 & 
         & 0.117 & 0.118 & 0.128 & 0.131 & 0.130 & {\scriptsize DyNet~\cite{shoshan2019dynet} } \\

         & 0.103 & 0.096 & \textbf{0.098} & \textbf{0.102} & 0.105 & 
         & 0.117 & 0.118 & 0.125 & 0.128 & 0.130 & {\scriptsize CFSNet~\cite{wang2019cfsnet} } \\
        
        \rowcolor{LightYellow} \cellcolor{White} 
         & \textbf{0.102} & \textbf{0.092} & \textbf{0.098} & 0.103 & \textbf{0.104} & \cellcolor{White} 
         & \textbf{0.116} & \textbf{0.111} & \textbf{0.115} & \textbf{0.123} & \textbf{0.129} & {\scriptsize Ours } \\
        
        \cmidrule(r){1-1} \cmidrule{2-6} \cmidrule{8-12} \cmidrule(l){13-13}
        
    \end{tabular}
\end{adjustbox}
\vspace{-0.25cm}
\caption{
Tuning $\times 4$ super-resolution. Increasing parameter $\omega_1$ promotes data fidelity, whereas increasing $\omega_2$ promotes perceptual quality. Our strategy is the best in almost all cases.
}
\vspace{-0.25cm}
\label{tbl:sr:gan}
\end{table}

%% file: figures/6_exp_sr_rec_gan.tex
\begin{figure}[t!]

    % trim=left bottom right top
    % \newcommand{\srincludegraphics}[1]{\includegraphics[trim=0px 20px 0px 10px, clip, width=0.3\columnwidth]{images/6_sr_rec_gan/#1}}
    \newcommand{\srincludegraphics}[1]{\includegraphics[width=0.3\columnwidth]{images/6_sr_rec_gan/camera_ready/#1}}
    
    \centering
    \scriptsize
    \setlength{\tabcolsep}{2pt}
    \begin{tabular}{c ccc}
         & \multicolumn{1}{l}{$\leftarrow$ Perceptual Quality}  & $(\mathcal{L}_{\text{rec}}, \mathcal{L}_{\text{gan}})$ & \multicolumn{1}{r}{Fidelity $\rightarrow$} \\[0.1cm]
         
         & $(0.00, 1.00)$ & $(0.50, 0.50)$ & $(1.00, 0.00)$ \\[0.1cm]
        
        \multirow{1}{*}[1.5cm]{\rotatebox[origin=c]{90}{{\scriptsize DyNet~\cite{shoshan2019dynet}}}} &
        \srincludegraphics{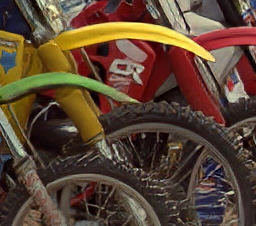} &
        \srincludegraphics{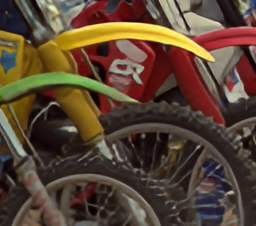} &
        \srincludegraphics{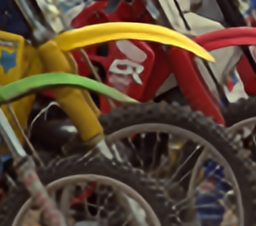} 
        
        \\
        
        \multirow{1}{*}[1cm]{\rotatebox[origin=c]{90}{{\scriptsize Ours}}} &
        \srincludegraphics{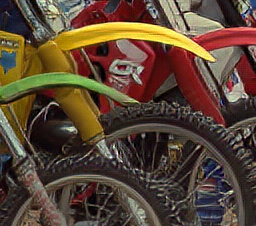} &
        \srincludegraphics{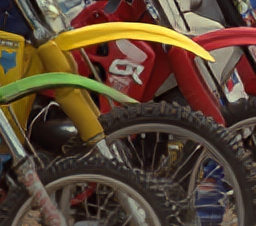} &
        \srincludegraphics{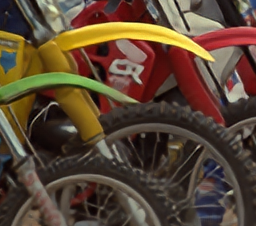}
    \end{tabular}
    \vspace{-0.25cm}
    \caption{
    Tuning $\times 4$ super-resolution perceptual quality on \mbox{image} 04 from the Kodak dataset~\cite{kodak}. Our method generates more detailed and perceptually pleasing results.
    }
    \label{fig:sr:gan}
\end{figure}

%% file: tables/5_exp_sr_rec_gan_sota.tex
\begin{table}[t]
\setlength{\tabcolsep}{3pt}

\begin{adjustbox}{scale=0.675,center}
    \centering
    \begin{tabular}{c cccc ccc l}
        & {\small \hspace{0.015cm}SAN\hspace{0.015cm}} 
        & {\small \hspace{0.015cm}HAN\hspace{0.015cm}} 
        & {\small \hspace{0.005cm}NLSA\hspace{0.005cm}} 
        & \multicolumn{3}{c}{{\small SwinIR~\cite{liang2021swinir} } }
        & \\

        & {\small \cite{dai2019san} }
        & {\small \cite{niu2020han} }
        & {\small \cite{mei2021nlsa} }
        & {\scriptsize Fixed}
        & \cellcolor{LightYellow} {\scriptsize $(1,0)$ }
        & \cellcolor{LightYellow} {\scriptsize $(0,1)$ } 
        & \\
    
        \cmidrule{1-1} \cmidrule(lr){2-2} \cmidrule(lr){3-3} \cmidrule(lr){4-4} \cmidrule{5-7} \cmidrule(l){8-8}
    
        % \textcolor{red}{}
        % \textcolor{blue}{}
        
        {\scriptsize PSNR}
        & 32.64 & 32.64 & 32.59 & \textbf{32.72}
        & \cellcolor{LightYellow} 32.57 & \cellcolor{LightYellow} 30.22 
        & \multirow{2}{*}[0cm]{\rotatebox[origin=c]{0}{{\small Set5~\cite{bevilacqua2012set5}}}} \\
        {\scriptsize SSIM}
        & 0.900 & 0.900 & 0.900 & \textbf{0.902} 
        & \cellcolor{LightYellow} 0.895 & \cellcolor{LightYellow} 0.838 & \\

        \cmidrule{1-1} \cmidrule(lr){2-2} \cmidrule(lr){3-3} \cmidrule(lr){4-4} \cmidrule{5-7} \cmidrule(l){8-8}
        
        {\scriptsize PSNR}
        & 28.92 & 28.90 & 28.87 & \textbf{28.94}
        & \cellcolor{LightYellow} 28.83 & \cellcolor{LightYellow} 26.27 
        & \multirow{2}{*}[0cm]{\rotatebox[origin=c]{0}{{\small Set14~\cite{zeyde2012set14}}}} \\
        {\scriptsize SSIM}
        & 0.789 & 0.789 & 0.789 & \textbf{0.791}
        & \cellcolor{LightYellow} 0.787 & \cellcolor{LightYellow} 0.701 & \\

        \cmidrule{1-1} \cmidrule(lr){2-2} \cmidrule(lr){3-3} \cmidrule(lr){4-4} \cmidrule{5-7} \cmidrule(l){8-8}
        
        {\scriptsize PSNR}
        & 27.78 & 27.80 & 27.78 & \textbf{27.83}
        & \cellcolor{LightYellow} 27.79 & \cellcolor{LightYellow} 25.05 
        & \multirow{2}{*}[0cm]{\rotatebox[origin=c]{0}{{\small BSD100~\cite{martin2001bsd68}}}} \\
        {\scriptsize SSIM}
        & 0.744 & 0.744 & 0.744 & \textbf{0.746} 
        & \cellcolor{LightYellow} 0.739 & \cellcolor{LightYellow} 0.647 & \\

        \cmidrule{1-1} \cmidrule(lr){2-2} \cmidrule(lr){3-3} \cmidrule(lr){4-4} \cmidrule{5-7} \cmidrule(l){8-8}

        {\scriptsize PSNR}
        & 26.79 & 26.85 & 26.96 & 27.07
        & \cellcolor{LightYellow} \textbf{27.10} & \cellcolor{LightYellow} 25.50 
        & \multirow{2}{*}[0cm]{\rotatebox[origin=c]{0}{{\small Urban100~\cite{huang2015single}}}} \\
        {\scriptsize SSIM}
        & 0.807 & 0.809 & 0.811 & \textbf{0.816} 
        & \cellcolor{LightYellow} 0.815 & \cellcolor{LightYellow} 0.764 & \\

        \cmidrule{1-1} \cmidrule(lr){2-2} \cmidrule(lr){3-3} \cmidrule(lr){4-4} \cmidrule{5-7} \cmidrule(l){8-8}
    
    \end{tabular}
\end{adjustbox}

\vspace{-0.2cm}
\caption{
SwinIR trained for tunable $\times 4$ super-resolution $(\mathcal{L}_{\text{rec}}$, $\mathcal{L}_{\text{gan}})$. Tuning for data fidelity $(1,0)$ results in performance comparable to the fixed baselines.
}
\label{tbl:sr:gan_sota}
\end{table}

%% file: figures/7_exp_style_mc_el_ky.tex
\begin{figure}[t!]

    % trim=left bottom right top
    % \newcommand{\styincludegraphics}[1]{\includegraphics[trim=0px 100px 0px 120px, clip, width=0.33\columnwidth]{images/7_style_mc_el_ky/#1}}
    \newcommand{\styincludegraphics}[1]{\includegraphics[width=0.315\columnwidth]{images/7_style_mc_el_ky/camera_ready/#1}}
    
    \centering
    \scriptsize
    \setlength{\tabcolsep}{2.5pt}
    \begin{tabular}{ccc}
         \multicolumn{3}{c}{ $(\mathcal{L}_{\text{mosaic}}, \mathcal{L}_{\text{edtaonisl}}, \mathcal{L}_{\text{kandinsky}})$ } \\[0.1cm]
         
        $(0.00, 0.00, 1.00)$ & $(0.00, 1.00, 0.00)$ & $(1.00, 0.00, 0.00)$ \\[-0.025cm]
        \styincludegraphics{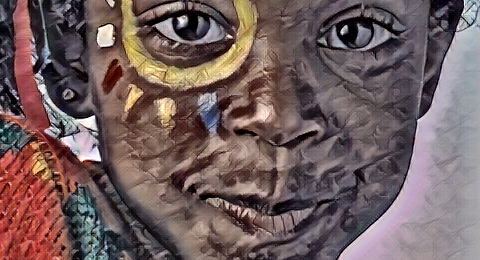} &
        \styincludegraphics{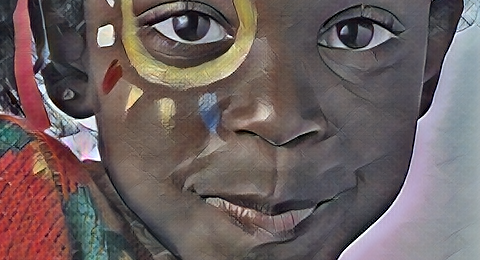} &
        \styincludegraphics{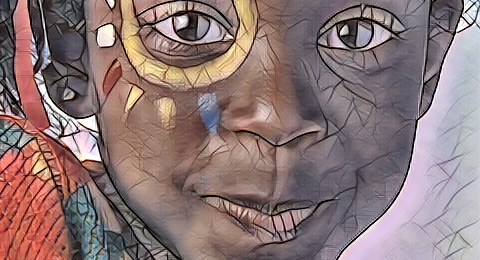} \\[-0.075cm]
        \styincludegraphics{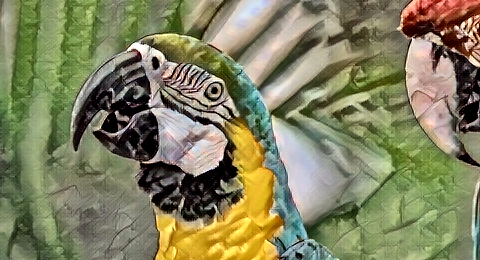} &
        \styincludegraphics{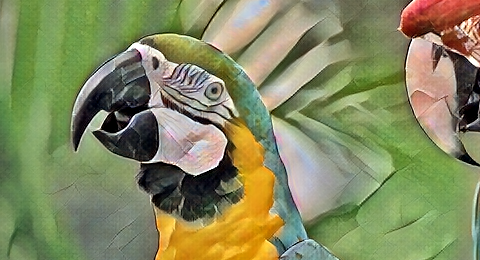} &
        \styincludegraphics{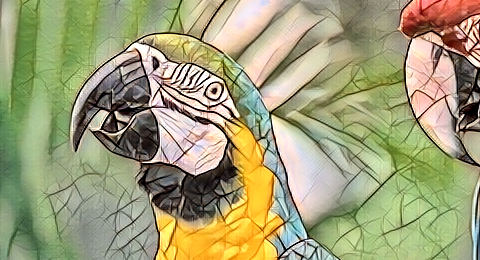} 
        
        \\[0.1cm]
        
        \styincludegraphics{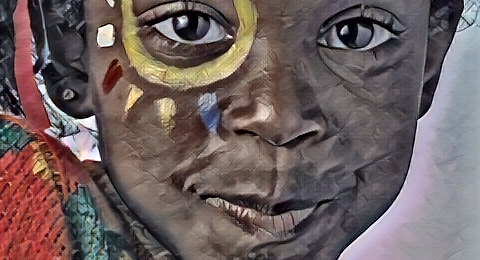} &
        \styincludegraphics{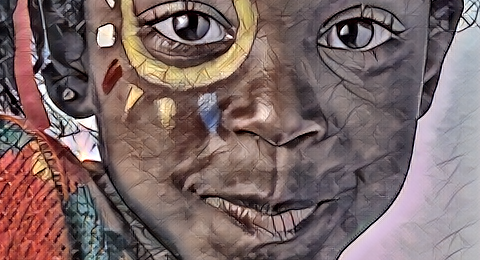} &
        \styincludegraphics{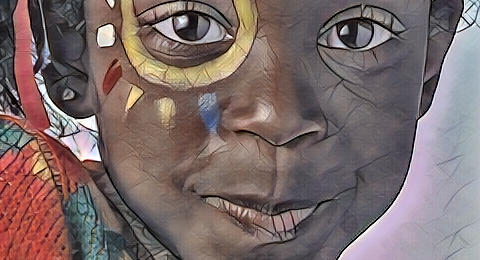} \\[-0.075cm]
        \styincludegraphics{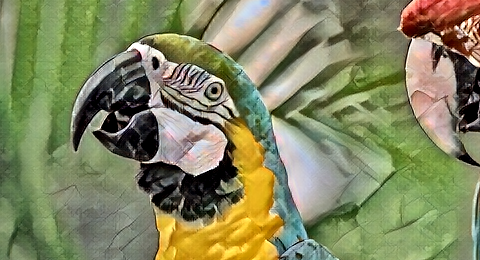} &
        \styincludegraphics{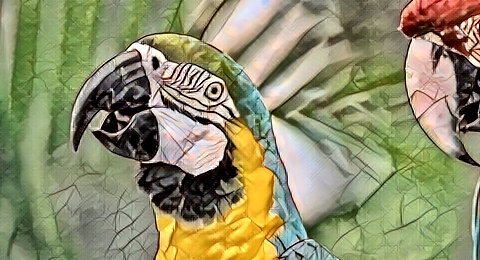} &
        \styincludegraphics{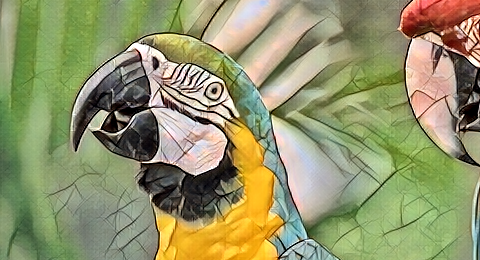} \\[-0.075cm]
        $(0.00, 0.50, 0.50)$ & $(0.50, 0.00, 0.50)$ & $(0.50, 0.50, 0.00)$
        
    \end{tabular}
    \vspace{-0.25cm}
    \caption{
    Our method is able to generate pleasing results for all style combinations by interacting with the parameters $(\omega_1, \omega_2, \omega_3)$ controlling \textit{Mosaic}, \textit{Edtaonisl}, and \textit{Kandinsky} style transfer on image 15 and 32 of the Kodak~\cite{kodak} dataset.
    }
    \label{fig:style}
    \vspace{-0.25cm}
\end{figure}

%% file: figures/8_tsne_blocks.tex
\begin{figure}[t]
\begin{center}
\setlength{\tabcolsep}{1pt}
\begin{tabular}{ccc}

    \multirow{1}{*}[0.9cm]{\rotatebox[origin=c]{90}{{\scriptsize Denoising}}} & 
    \includegraphics[trim=40px 40px 10px 10px,clip,width=0.9\columnwidth]{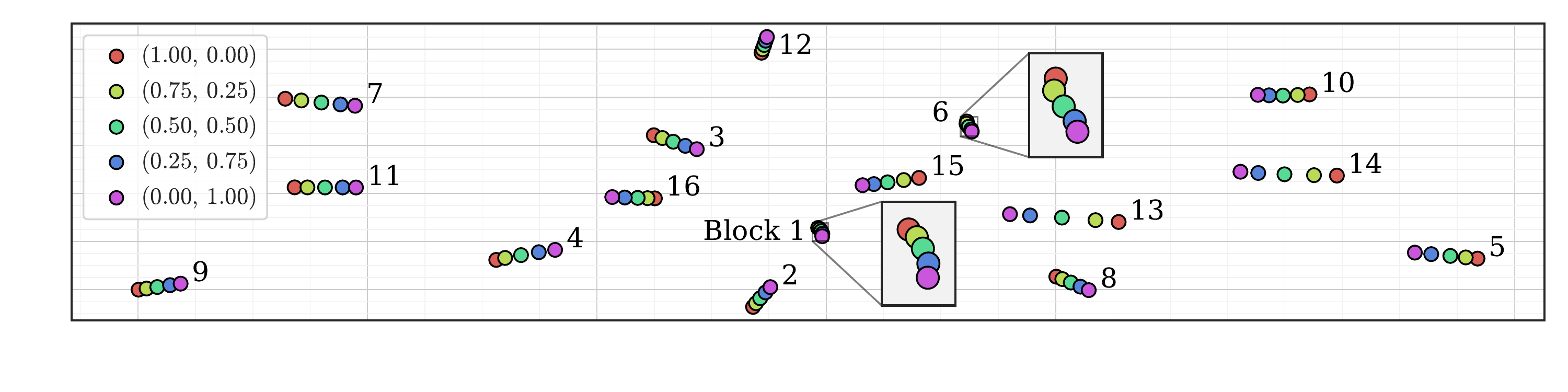} & 
    \multirow{1}{*}[0.8cm]{\rotatebox[origin=c]{90}{{\scriptsize Table~\ref{tbl:dn:strength} }}} \\
    
    \multirow{1}{*}[1.0cm]{\rotatebox[origin=c]{90}{{\scriptsize Super-Res.}}} & 
    \includegraphics[trim=40px 40px 10px 10px,clip,width=0.9\columnwidth]{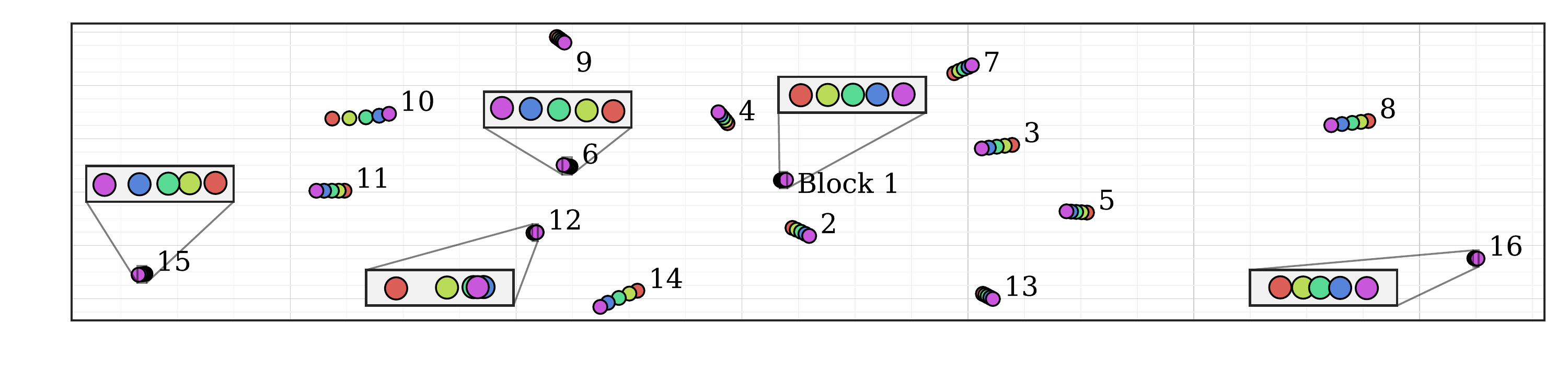} & 
    \multirow{1}{*}[0.8cm]{\rotatebox[origin=c]{90}{{\scriptsize Table~\ref{tbl:sr:gan}}}}
    
\end{tabular}
\end{center}
\vspace{-0.6cm}
\caption{
t-SNE of the tuned kernels in the $16$ residual blocks of our tunable denoising and super-resolution networks. The kernels span quasi-linear manifolds of different lengths and orientations between the two objectives $(0.00, 1.00)$ and $(1.00, 0.00)$. 
}
\label{fig:tsne:resblocks}
\end{figure}

%% file: figures/9_kernels.tex
\begin{figure}[t]
\begin{center}

    \setlength{\tabcolsep}{2pt}
    \renewcommand{\arraystretch}{0.6}
    \begin{tabular}{c | ccc | ccc}

    % {\tiny Layer}
    & \multicolumn{3}{c|}{{\tiny Denoising}} 
    & \multicolumn{3}{c}{{\tiny Super-Resolution}} \\

    \hline 
    
    \multirow{1}{*}[0.175cm]{{\tiny Input}} 
    & \multicolumn{3}{c|}{\includegraphics[width=0.42\columnwidth]{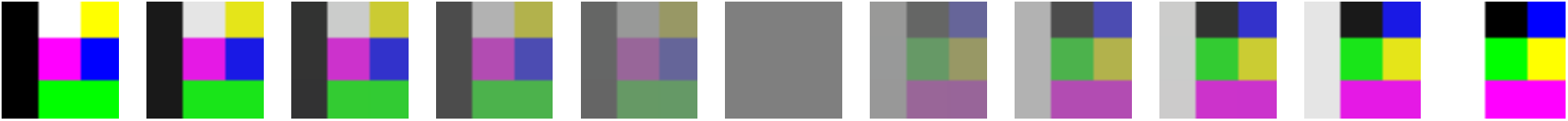}} 
    & \multicolumn{3}{c}{\includegraphics[width=0.42\columnwidth]{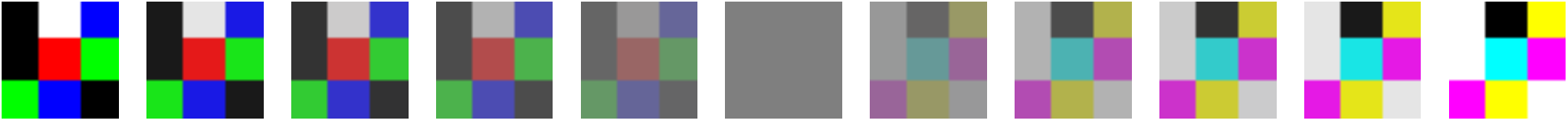}} \\
    
    \multirow{1}{*}[0.2cm]{{\tiny Output}} 
    & \multicolumn{3}{c|}{\includegraphics[width=0.42\columnwidth]{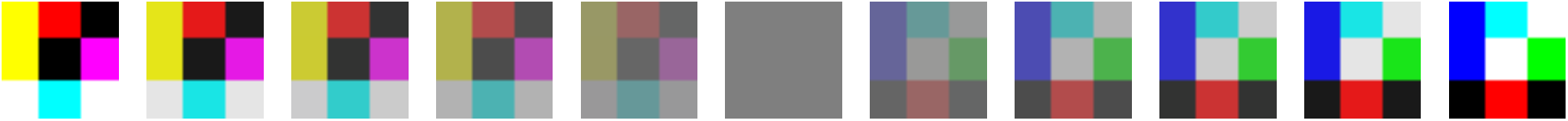}} 
    & \multicolumn{3}{c}{\includegraphics[width=0.42\columnwidth]{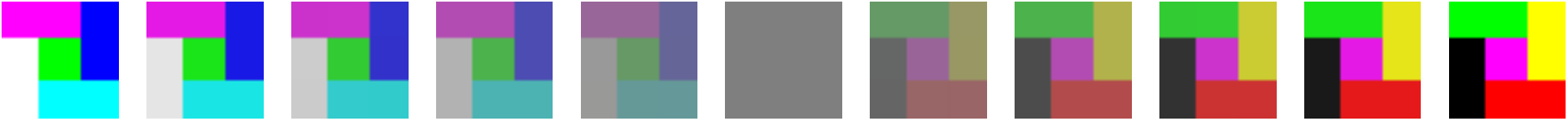}} \\[-0.1cm]

    \hline
    
    {\tiny $\omega_1$}
    & \multicolumn{1}{l }{                {\tiny $0.0$}}
    & \multicolumn{1}{c }{\hspace{1.117cm}{\tiny $0.5$}}
    & \multicolumn{1}{r|}{                {\tiny $1.0$}}
    
    & \multicolumn{1}{l }{                {\tiny $0.0$}}
    & \multicolumn{1}{c }{\hspace{1.117cm}{\tiny $0.5$}}
    & \multicolumn{1}{r }{                {\tiny $1.0$}} \\[-0.05cm]

    {\tiny $\omega_2$}
    & \multicolumn{1}{l }{                {\tiny $1.0$}}
    & \multicolumn{1}{c }{\hspace{1.117cm}{\tiny $0.5$}}
    & \multicolumn{1}{r|}{                {\tiny $0.0$}}
    
    & \multicolumn{1}{l }{                {\tiny $1.0$}}
    & \multicolumn{1}{c }{\hspace{1.117cm}{\tiny $0.5$}}
    & \multicolumn{1}{r }{                {\tiny $0.0$}} \\
    
    % {\tiny $\boldsymbol{\omega}$}
    % & \multicolumn{1}{l}{{\tiny $(1,0)$ \hspace{0.0cm} $\cdots$}} 
    % & \multicolumn{1}{c}{{\tiny \hspace{0.0cm} $(0.5,0.5)$}} 
    % & \multicolumn{1}{r|}{{\tiny $\cdots$ \hspace{0.0cm} $(0,1)$}}
    % & \multicolumn{1}{l}{{\tiny $(1,0)$ \hspace{0.0cm} $\cdots$}} 
    % & \multicolumn{1}{c}{{\tiny \hspace{0.0cm} $(0.5,0.5)$}} 
    % & \multicolumn{1}{r}{{\tiny $\cdots$ \hspace{0.0cm} $(0,1)$}} \\
    
    \end{tabular}
    
\end{center}
\vspace{-0.6cm}
\caption{Input and output kernels tuned with uniform parameters reveal a linear behavior while transitioning between objectives.}
\label{fig:pca:ker}
\end{figure}